# Incremental computation of the value of perfect information in stepwise-decomposable influence diagrams

(Nevin) Lianwen Zhang, Runping Qi, and David Poole
Department of Computer Science
University of British Columbia
Vancouver, B.C., Canada
{lzhang, qi, poole}@cs.ubc.ca

## Abstract

To determine the value of perfect information in an influence diagram, one needs first to modify the diagram to reflect the change in information availability, and then to compute the optimal expected values of both the original diagram and the modified diagram. The value of perfect information is the difference between the two optimal expected values. This paper is about how to speed up the computation of the optimal expected value of the modified diagram by making use of the intermediate computation results obtained when computing the optimal expected value of the original diagram.

## 1  INTRODUCTION

The concept of the value of perfect information is very useful in gaining insights about decision problems. Matheson (1990) has demonstrated that influence diagrams provide a more suitable paradigm to address the issue than decision trees. This paper is concerned with the problem of computing the value of perfect information in influence diagrams.

An influence diagram is a graphical representation of a particular decision problem (Howard and Matheson 1984). It is an acyclic directed graph with three types of nodes: decision nodes, random nodes and value nodes. The influence diagram in Fig. 1 represents an extension to the well-known oil wildcatter problem (Raiffa 1968, Shachter 1986). The decision nodes are depicted as rectangles, random nodes as ellipses, and value nodes as diamonds.

The total value of the diagram is `oil-sales` minus the sum of `test-cost`, `drill-cost`, and `sale-cost`. The expectation of the total value depends on the decisions made. The *optimal expected value of the diagram* is defined to be the maximum of the expected total values.

There are arcs from `test` and `test-result` to

`drill`. This means that the `drill` decision is to be made knowing the type of `test` performed and the `test-result`. There is no arc from `market-information` to `drill`. This means that `market-information` (at the time when the `oil-sale-policy` is to be made) is not available at the time the `drill` decision is to be made.

To reduce risks, the oil wildcatter may choose to, before making the `drill` decision, hire an expert to predicate `market-information` at the time the `oil-sale-policy` is to be made. This operation may be expensive. So, before making up his mind, the oil wildcatter may wish to determine the value of the expert's predications. The value of perfect information on `market-information` serves as an upper bound for the value of the predications. Specifically, it is defined as follows. Modify the diagram by adding an arc from `market-information` to `drill`. The *value of perfect information* on `market-information` is the difference between the optimal expected value of the modified diagram and that of the original diagram.

A straightforward method of computing the value of perfect information is to exactly follow the definition. That is to respectively compute the optimal expected values of the original influence diagram and of the modified diagram, and then figure out the difference. This paper shows that one can do better than that. Since the original and the modified diagrams differ very little, there must be computation overlaps in the processes of evaluating them. By avoiding those overlaps, one can speed up computing the optimal ex-pected value of the modified diagram. This is especially interesting if one wants to assess the value of perfect information for a number of cases.

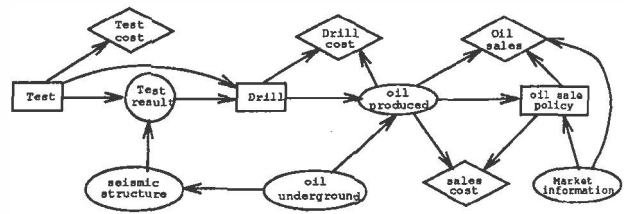

Figure 1: The influence diagram for an extension of the oil wildcatter problem.

The exposition will be carried out in the terms of stepwise-decomposable influence diagrams, which is reviewed in section 2. Section 3 introduces the concept of influence diagram condensation. Section 4 shows how to use this concept to uncover the computation overlaps mentioned in the previous paragraph. The paper concludes at section 5.

## 2   STEPWISE-DECOMPOSABLE INFLUENCE DIAGRAMS

Stepwise-decomposable influence diagrams (SDID's) were first introduced by Zhang and Poole (1992) as a generalization to the traditional notion of influence diagrams. Better references in this regard are Zhang, Qi and Poole (1993b) and Zhang (1993). This section reviews the concept of SDID's.

### 2.1   THE PATH TO SDID'S

Traditionally, there are five constraints imposed on influence diagrams:

1. *Acyclicity*, which requires that there be no directed loops in influence diagrams;

2. *Regularity*, which requires that there be a total ordering among all the decision nodes;

3. *The no-forgetting constraint*, which requires that any decision node and its parents be parents to all subsequent decision nodes;

4. *The single-value-node constraint*, which requires that there be only one value node; and

5. *The leaf-value-node constraint*, which requires that the value node have no children.

Influence diagrams that satisfy all the five constraints will be referred to as *no-forgetting influence diagrams*.

We propose to lift constraints 2-4 and to develop a general theory of influence diagrams starting with constraints 1 and 5 only.

There are several advantages to lift constraints 2-4. For instance, by lifting the no-forgetting constraint we are able to, anmong other things, represent the facts that some decision nodes are conditionally independent of certain pieces of information. In the extended oil wildcatter problem (Fig. 1), it is reasonable to assume that the decision `oil-sale-policy` is independent of information on `test-result` given the quality and quantity of `oil-produced`. This piece of knowledge can not be represented if the no-forgetting constraint is enforced. The reader is referred to Zhang (1993), Zhang, Qi and Poole (1993) for other rationale for lifting constraints 2-4.

Note: Traditionally, arcs into decision nodes are interpreted as indications of information availability. Now that the no-forgetting constraint has been lifted, those arcs need to be re-interpreted as indication of both information availability and dependency. More explicitly, the lack of an arc from a node $c$ to a decision node $d$ no longer implies that information $c$ is not observed when making decision $d$. It may as well mean that $d$ is independent of $c$ given the parents of $d$.

Acyclicity and the leaf-node constraint together define a very general concept of influence diagrams. One theme of Zhang (1993) is to identify subclasses of influence diagrams with various computational properties. One important property for an influence diagram to possess is the so-called stepwise-solvability, which says that the diagram can be evaluated by considering one decision node at a time. If an influence diagram is not stepwise-solvable, then one needs to simultaneously consider several, even all the decision nodes, which usually tends to be computationally expensive.

When an influence diagram is stepwise-solvable? The answer: when it is stepwise-decomposable. It can be shown that a stepwise-decomposable influence diagram can be evaluated not only by considering one decision node at a time, but also by considering one section of the diagram at a time (Zhang, Qi and Poole 1993). It can also be shown that an influence diagram is stepwise-solvable only when it is stepwise-decomposable (Zhang 1993).

In the rest of this section, we define stepwise-decomposable influence diagrams.

### 2.2   INFLUENCE DIAGRAMS

An *influence diagram* is an acyclic directed graph consisting of a set of random nodes $C$, a set of decision nodes $D$, and a set of value nodes $U$. The value nodes have no children. A random node $c$ represents an uncertain quantity whose value is determined according to a given conditional probability distribution $P(c|\pi_c)$, where $\pi_c$ stands for the set of the parents of $c$. A value node $v$ represent one portion of the decision maker's utilities, which is characterized by a *value function* $f_v$.

Let $I$ be an influence diagram. For any node $x$ in $I$, let $\pi_x$ denote the set of the parents of $x$. Let $\Omega_x$ denote the *frame* of $x$, i.e the set of possible values of $x$. For any set $J$ of nodes, let $\Omega_J = \prod_{x \in J} \Omega_x$.

Let $d_1, \ldots, d_k$ be all the decision nodes in $I$. For a decision node $d_i$, a mapping $\delta_i : \Omega_{\pi_{d_i}} \to \Omega_{d_i}$ is called a *decision function* for $d_i$. The set of all the decision functions for $d_i$, denoted by $\Delta_i$, is called the *decision function space* for $d_i$. The Cartesian product of the decision function spaces for all the decision nodes is called the *policy space* of $I$. We denote it by $\Delta$.

Given a policy $\delta = (\delta_1, \ldots, \delta_k) \in \Delta$ for $I$, a probability $P_\delta$ can be defined over the random nodes and the decision nodes as follows:

$$P_\delta(C, D) = \prod_{c \in C} P(c|\pi_c) \prod_{i=1}^{k} P_{\delta_i}(d_i|\pi_{d_i}), \qquad (1)$$

where $P(c|\pi_c)$ is given in the specification of the influence diagram, while $P_{\delta_i}(d_i|\pi_{d_i})$ is given by $\delta_i$ as follows:

$$P_{\delta_i}(d_i|\pi_{d_i}) = \begin{cases} 1 & \text{if } \delta_i'(\pi_{d_i}) = d_i, \\ 0 & \text{otherwise} \end{cases} \qquad (2)$$

For any value node $v$, $\pi_v$ must consist of only decision and value nodes, since value nodes do not have children. Hence, we can talk about $P_\delta(\pi_v)$. The *expectation of the value node $v$ under $P_\delta$*, denoted by $E_\delta[v]$, is defined as follows:

$$E_\delta[v] = \sum_{\pi_v} P_\delta(\pi_v) f_v(\pi_v).$$

The summation of the expectations of all the value nodes is called the *value* of $I$ under the policy $\delta$, We denote this denoted by $E_\delta[I]$. The maximum of $E_\delta[I]$ over all the possible policies $\delta$ is the *optimal expected value* of $I$. An *optimal policy* is a policy that achieves the optimal expected value. To *evaluate* an influence diagram is to determine its optimal expected value and to find an optimal policy.

An influence is *regular* if there exists a total ordering among all the decision nodes. Even though all our results in this paper readily generalizes to influence diagrams which are not necessarily regular, we shall limit the exposition only to regular influence diagrams for the sake of simplicity.

## 2.3    STEPWISE-DECOMPOSABLE INFLUENCE DIAGRAMS

To introduce stepwise-decomposable influence diagrams (SDID), we need the concepts of moral graph and of m-separation. Let $G$ be a directed graph. The *moral graph* $G$ is an undirected graph $m(G)$ with the same vertex set as $G$ such that there is an edge between two vertices in $m(G)$ if and only if either there is an arc between them in $G$ or they share a common child in $G$. The term moral graph was chosen because two nodes with a common child are "married" into an edge (Lauritzen and Spiegelhalter 1988).

In an undirected graph, two nodes $x$ and $y$ are *separated* by a set of nodes $A$ if every path connecting them contains at least one node in $A$. In a directed graph $G$, $x$ and $y$ are *m-separated* by $A$ if they are separated by $A$ in the moral graph $m(G)$. One implication of this definition is that $A$ m-separates every node in $A$ from any node outside $A$.

For any decision node $d$ of $I$, the *downstream* of $\pi_d$ is the set of nodes that are not m-separated from $d$ by $\pi_d$. The *upstream* of $\pi_d$ is that set of nodes outside $\pi_d$ that are m-separated from $d$ by $\pi_d$.

A regular influence diagram is *stepwise-decomposable* if for any decision node $d$, none of the decision nodes that precede $d$ are in the downstream of $\pi_d$.

The influence diagram in Fig. 1 is a SDID. No-forgetting influence diagrams are SDID's.

One desirable property of SDID's is that they are stepwise-solvable. As an example, consider the SDID in Fig. 1. One can first compute an optimal policy for `oil-sale-policy` in the part of the diagram that lies to the right of `oil-produced` with `oil-produced` included, and then find an optimal policy for `drill`, and then for `test`. The optimal expected value of the diagram obtained as a by-product of computing an optimal policy for `test`. See Zhang, Qi and Poole (1993b) for details.

## 3    CONDENSING SDID'S

This section presents a two-stage approach for evaluating SDID's. In the first stage, a SDID is "condensed" into a Markov decision process (Denardo 1982). This involves two types of operations: the operation of computing conditional probabilities and the operation of summing up several functions. In the second stage, the condensed SDID is evaluated by the various algorithms (Qi 1993, Qi and Poole 1993).

This two-stage approach is interesting because it allows easy implementation of influence diagrams on top of a system for Bayesian network computations (Zhang 1993). The approach is also of fundamental significance to the current paper, as the reader will see in Section 4.

This approach has been developed from a similar approach in terms of decision graphs (Qi 1993, Zhang, Qi and Poole 1993a). In the rset of this paper, we shall concentrate on the first stage, i.e. condensation. Let us begin with smoothness in SDID's.

## 3.1    SMOOTHNESS IN SDID'S

An influence diagram is *smooth at a decision node $d$* if there is no arcs from the downstream of $\pi_d$ to $\pi_d$. If an influence diagram is smooth at all the decision nodes, we say that the diagram is *smooth*.

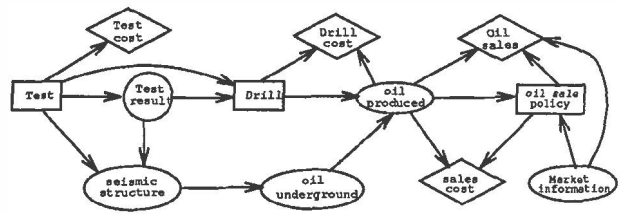

Figure 2: The influence diagram in Fig. 1 after smoothing.

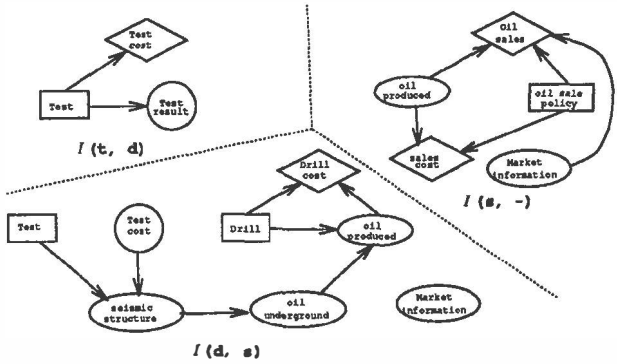

Figure 3: The sections of the SDID in Fig. 2.

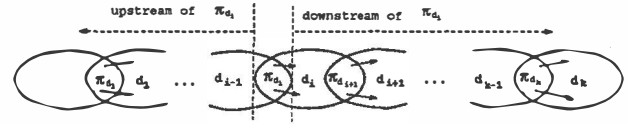

Figure 4: An abstract view of a smooth regular SDID.

SDID's may be not smooth. For example, the SDID in Fig. 1 is not smooth at the decision node `drill`. The arc from `seismic-structure` to `test-result` is from the downstream of $\pi_{\text{drill}}$ to $\pi_{\text{drill}}$.

Two influence diagrams are *strongly equivalent* if they have the same set of nodes, the same optimal policies, and the same optimal expected value. A non-smooth SDID can always be transformed, by a series of arc reversals (Shachter 1986), into a strongly equivalent smooth SDID (Zhang, Qi, and Poole 1993b). For example, the SDID in Fig. 1 can be transformed into a strongly equivalent SDID whose underlying graphical structure is shown in Fig. 2. This SDID is smooth.

From now on, we shall only be talking about smooth SDID's.

## 3.2   SECTIONS IN SDID'S

The concept of sections in SDID is a prerequisite for the concept of condensation.

Let $I$ be a smooth regular SDID. Let $d_1$, $d_2$, ..., $d_k$ be the decision nodes. Since $I$ is regular, there is a total ordering among the decision nodes. Let the total ordering be as indicated by the subscriptions of the decision nodes. As a consequence, we have that $d_i$ precedes $d_{i+1}$, and there is no other decision node $d$ such that $d_i$ precedes $d$ and $d$ precedes $d_{i+1}$.

For any $i \in \{1, 2, \ldots, k-1\}$, the *section of I from $\pi_{d_i}$ to $\pi_{d_{i+1}}$*, denoted by $I(d_i, d_{i+1})$, is the subnetwork of $I$ that consists of the following nodes:

1. the nodes in $\pi_{d_i} \cup \pi_{d_{i+1}}$,
2. the nodes that are in both the downstream of $\pi_{d_i}$ and in the upstream of $\pi_{d_{i+1}}$,

The graphical connections among the nodes remain the same as in $I$ except that all the arcs among the nodes in $\pi_{d_i} \cup \{d_i\}$ are removed.

The *initial section $I(-, d_1)$* consists of the nodes in $\pi_{d_1}$ and the nodes in the upstream of $\pi_{d_1}$. It consists of only random and value nodes.

The *terminal section $I(d_k, -)$* consists of the nodes in the $\pi_{d_k}$ and the nodes in the downstream of $\pi_{d_k}$.

The nodes in a section that lie outside $\pi_{d_i} \cup \{d_i\}$ are either random nodes or value nodes. Their conditional probabilities and value functions are the same as those in $I$. The nodes in $\pi_{d_i} \cup \{d_i\}$ are either decision nodes or random nodes. There are no conditional probabilities are associated with these nodes.

Let us temporarily denote the SDID in Fig. 2 by $I$. Let us denote the variables `test` by `t`, `drill` by `d`, `oil-sale-policy` by `s`, `drill-cost` by `dc`, `test-result` by `tr`, `oil-produced` by `op`, and `market-information` by `mi`.

There are four sections in this SDID: $I(-, t)$, $I(t, d)$, $I(d, s)$, and $I(s, -)$. The initial section $I(-, t)$ is empty. All the other sections are shown in Fig. 3.

The concept of sections provides us with a perspective of viewing smooth regular SDID's. A smooth regular SDID $I$ can be thought of as consisting of a chain sections $I(-, d_1)$, $I(d_1, d_2)$, ..., $I(d_{k-1}, d_k)$, and $I(d_k, -)$. Two neighboring sections $I(d_{i-1}, d_i)$ and $I(d_i, d_{i+1})$ share the nodes in $\pi_{d_i}$, which m-separate the other nodes in $I(d_{i-1}, d_i)$ from all the other nodes $I(d_i, d_{i+1})$. Fig. 4 shows this abstract view of a smooth regular SDID.

In the extended oil wildcatter example as shown in Fig. 3, the sections $I(t, d)$ and $I(d, s)$ share the nodes `test` and `test-result`, and the sections $I(d, s)$ and $I(s-)$ share the nodes `oil-produced` and `market-information`.

## 3.3   CONDITIONAL PROBABILITIES AND LOCAL VALUES IN THE SECTIONS

In the section $I(d_i, d_{i+1})$, there is no decision node outside $\pi_{d_i} \cup \{d_i\}$. The value nodes are at leaves by definition. So, one is able to compute the conditional probability $P_{I(d_i, d_{i+1})}(\pi_{d_{i+1}} | \pi_{d_i}, d_i)$ of the nodes in $\pi_{d_{i+1}}$ given the nodes in $\pi_{d_i}$ and $d_i$. We shall refer to this probability as the *conditional probability of $\pi_{d_{i+1}}$ given $\pi_{d_i}$ and $d_i$ in $I$*.

In the initial section $I(-, d_1)$, one can compute the probability $P_{I(-, d_1)}(\pi_{d_1})$. We shall refer to this probability as the *prior probability $\pi_1$ in $I$*.

For a value node $v_j$ in $I(d_i, d_{i+1})$, one can compute conditional probability $P_{I(d_i, d_{i+1})}(\pi_{v_j} | \pi_{d_i}, d_i)$. Define

a function $f_{v_j}^c : \Omega_{\pi_{d_i}} \times \Omega_{d_i} \to \mathcal{R}$ by

$$f_{v_j}^c(\pi_{d_i}, d_i) = \sum_{\pi_{v_j} - (\pi_{d_i} \cup \{d_i\})} P_{I(d_i, d_{i+1})}(\pi_{v_j} | \pi_{d_i}, d_i) f_{v_j}(\pi_{v_j}) \quad (3)$$

where $f_{v_j}$ is the value function of $v_j$ in $I$.

Let $v_1, \ldots, v_m$ be all the value nodes in the section $I(d_i, d_{i+1})$. The *local value function* $f_i : \Omega_{\pi_{d_i}} \times \Omega_{d_i} \to \mathcal{R}$ of the section $I(d_i, d_{i+1})$ is defined by

$$f_i(\pi_{d_i}, d_i) = \sum_{j=1}^m f_{v_j}^c(\pi_{d_i}, d_i). \quad (4)$$

When there are no value node in the section, then $f_i$ is defined to be the constant 0.

We can also define the local value "function" for the initial section, which is not really a function, but just a constant. We shall denote this constant by $f_0$.

## 3.4  CONDENSATION

Intuitively, condensing a smooth regular SDID $I$ means to do the following in each section $I(d_i, d_{i+1})$ of $I$: (1) getting rid of all the random nodes that are neither in the $\pi_{d_i}$ nor in $\pi_{d_{i+1}}$, (2) combining all the value nodes into one single value node $v_i^c$, and (3) collecting the nodes in $\pi_{d_i}$ into one compound variable $x_i$. This results in a Markov decision process.

Now the formal definition. The *condensation* of $I$, denoted by $I^c$, is defined as follows:

1. It consists of the following nodes:
   - Random nodes $x_i$ $(0 \le i \le k)$, where $x_i$ is the compound variable consists of all the nodes in $\pi_{d_i}$ when $\pi_{d_i} \ne \emptyset$. When $\pi_{d_i} = \emptyset$ or when $i = 0$, $x_i$ is a degenerated variable that has only one possible value, say, $\diamond^1$.
   - The same decision nodes $d_i$ $(1 \le i \le k)$ as in $I$; and
   - Value nodes $v_i^c$ $(0 \le i \le k)$,

2. The graphical connections among the nodes are as follows:
   - For any $i \in \{2, 3, \ldots, k\}$, there are two arcs converging at $x_i$, one from $x_{i-1}$ and the other from $d_{i-1}$.
   - For any $i \in \{1, 2, \ldots, k\}$, there is an arc from $x_i$ to $d_i$.
   - For any $i \in \{1, 2, \ldots, k\}$, there are two arcs converging at $v_i^c$, one from $x_i$ and the other from $d_i$.

---

[1]The presence of the node $x_0$ makes the picture ugly. But we need it for two reasons. First, there may be value nodes in the initial section. Second, we want to be able to talk about the condensation of an influence diagram that contains no decision nodes.

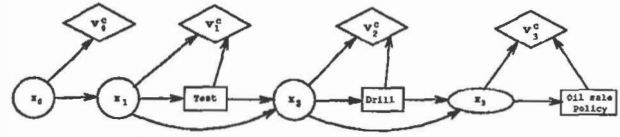

Figure 5: The condensation of the SDID in Fig. 2.

- There are two arcs emitting from $x_0$, one to $v_0^c$ and the other to $x_1$.

3. The conditional probabilities and value functions are as follows:
   - The conditional probability $P^c(x_{i+1} | x_i, d_i)$ $(i \in \{1, \ldots, k-1\})$ is defined to be $P_{I(d_{i+1}, d_i)}(\pi_{d_{i+1}} | \pi_{d_i}, d_i)$;
   - The conditional probability $P^c(x_1 | x_0 = \diamond)$ is defined to be $P_{I(-, d_1)}(\pi_{d_1})$, and the probability $P^c(x_0)$ is trivially defined since $x_0$ takes only one value $\diamond$;
   - The value function $f_{v_i^c}$ for $v_i^c$ $(i \in \{0, 1, \ldots, k\})$ is defined to be $f_i$.

Fig. 5 depicts the condensation of the SDID shown Fig. 2. Since **test** has no parent, $x_1$ is a degenerated variable. The variable $x_2$ stands for the compound variable consisting of **test** and **test-cost**, and $x_3$ stands for the compound variable consisting of **oil-produced** and **oil-market**. The conditional probability $P^c(x_3 | x_2, d)$, for instance, is the conditional probability $P_{I(d, s)}(\text{op}, \text{mi} | t, \text{tr}, d)$ of **op** and **mi** given **t**, **tr**, and **d** in the section $I(d, s)$.

The value function $f_{v_1^c}$ for the value node $v_1^c$ is a representation of **test-cost**, $f_{v_2^c}$ is a representation of **drill-cost**, and $f_{v_3^c}$ is a representation of **oil-produced** and **sale-cost**.

There is no value node in the initial section. So $f_{v_0^c}$ is the constant 0. The node is kept only for uniformity.

Two decision networks are *equivalent* if they have the same optimal value and share the same optimal policies. The following theorem is proved in Zhang (1993).

**Theorem 1** *A smooth regular SDID is equivalent to its condensation.*

To end this section, we would like to echo what we said at the beginning of the section. The process of condensing a SDID only involves two types of operations: the operation of computing conditional probabilities and the operation summing up functions (see subsection 3.3). The latter is straightforward. The formmer can be carried out by any well established Bayesian network evaluation algorithm. One advantage of the concept of condensation is that it leads to a simple way of implementing influence diagrams on top of a system for Bayesian network computation.

# 4   COMPUTING THE VALUE OF PERFECT INFORMATION

Let $I$ be a regular SDID. Let $d_s$ and $c$ respectively be a decision node and a random node in $I$, such that there is no arc from $c$ to $d_s$ in $I$. Let $I'$ be the diagram obtained from $I$ by adding arcs from $c$ to $d_s$ and to all the subsequent decision nodes. If there is no direct cycles in $I'^2$, then $I'$ is again a regular SDID. In such a case, *the value of perfect information on $c$ at $d_s$ in $I$* is defined to be the difference between the optimal expected value of $I'$ and that of $I$.

In the following, we shall use $\pi'_d$ to denote the set of parents of $d$ in $I'$.

To determine the value of perfect information on $c$ at $d_s$, one needs to compute the optimal expected values of both $I$ and $I'$. To this end, we adopt the two stage approach described in the previous section, i.e we first compute the condensations of $I$ and $I'$, and then evaluate the condensations respectively. An advantage of this approach is that it can make use of information stored in the condensation of $I$ in computing the condensation of $I'$. More explicitly, for each section $I(d_i, d_{i+1})$ of $I$, the conditional probability $P_{I(d_i, d_{i+1})}(\pi_{d_{i+1}} | \pi_{d_i}, d_i)$ and the local value function $f_i$ are computed and stored in the condensation of $I$. This paper seeks to make use of this conditional probability and this local value function in computing the conditional probability $P_{I'(d_i, d_{i+1})}(\pi'_{d_{i+1}} | \pi'_{d_i}, d_i)$ and the local value function $f'_i$ of the corresponding section $I'(d_i, d_{i+1})$ of $I'$.

To see an example, 'let $I$ be the SDID shown in Fig. 2. Consider the value of perfect information on **market-information** at **drill**. In this case, $I'$ is the same as $I$ expect for the arc from **market-information** to **drill**. The section $I'(\mathbf{s}, -)$ is the same as $I(\mathbf{s}, -)$. Thus, when computing the condensation of $I'$, the conditional probability and the local value function for this section can simply be retrieved from the condensation of $I$.

The section $I'(\mathbf{t}, \mathbf{d})$ is the same as $I(\mathbf{t}, \mathbf{d})$ except that it con­tains one extra random node **market-information**. The node **market-information** is isolated in $I'(\mathbf{t}, \mathbf{d})$. In the condensation of $I'$, one needs $P_{I'(\mathbf{t}, \mathbf{d})}(\mathtt{mi}, \mathtt{tr} | \mathtt{t})$. This can be computed by

$$P_{I'(\mathbf{t}, \mathbf{d})}(\mathtt{mi}, \mathtt{tr} | \mathtt{t}) = P_{I(\mathbf{t}, \mathbf{d})}(\mathtt{tr} | \mathtt{t}) P(\mathtt{mi}),$$

where $P(\mathtt{mi})$ is given in the specification of the diagram and $P_{I(\mathbf{t}, \mathbf{d})}(\mathtt{tr} | \mathtt{t})$ can be retrieved from the condensation of $I$.

The section $I'(\mathbf{d}, \mathbf{s})$ is also the same as $I(\mathbf{d}, \mathbf{s})$. How­ever, the decision nodes **drill** has one more parent, namely **market-information** in $I'$ than in $I$. Thus, to obtain the condensation of $I'$, one needs the conditional probability $P_{I'(\mathbf{d}, \mathbf{s})}(\mathtt{op}, \mathtt{mi} | \mathtt{t}, \mathtt{tr}, \mathtt{d}, \mathtt{mi})$. In the condensation of $I$, one has $P_{I(\mathbf{d}, \mathbf{s})}(\mathtt{op}, \mathtt{mi} | \mathtt{t}, \mathtt{tr}, \mathtt{d})$. The nice thing is that one can easily compute $P_{I'(\mathbf{d}, \mathbf{s})}(\mathtt{op}, \mathtt{mi} | \mathtt{t}, \mathtt{tr}, \mathtt{d}, \mathtt{mi})$ from $P_{I'(\mathbf{d}, \mathbf{s})}(\mathtt{op}, \mathtt{mi} | \mathtt{t}, \mathtt{tr}, \mathtt{d})$, which is the same as $P_{I(\mathbf{d}, \mathbf{s})}(\mathtt{op}, \mathtt{mi} | \mathtt{t}, \mathtt{tr}, \mathtt{d})$, which in turn can be retrieved from the condensation of $I$.

To summarize, it takes very little computation to ob­tain the condensation of $I'$ from the condensation of $I$. The rest of this section is to show that the same can be true for many other cases. We shall do this case by case. But first, some preparations.

## 4.1   REMOVABLE ARCS

A random node can be in more than one section. In the oil wildcatter example, **market-information** is in both the section $I(\mathbf{d}, \mathbf{s})$ and the section $I(\mathbf{s}, -)$. Let $d_t$ be the last decision node such that $c$ is in the section $I(d_{t-1}, d_t)$ (remember that there is a total ordering among the decision nodes).

The reader is advised to pay close attention to the definition of $d_t$ and the definition of $d_s$ (given at the beginning of this section), ince we shall use them fre­quently in the rest of the paper.

It follows from a result of Zhang and Poole (1992) that in $I'$ the arcs from $c$ to the decision nodes subsequent to $d_t$, i.e to the decision nodes $d_{t+1}, \ldots, d_k$ are remov­able, in the sense that the removal of those arcs results in an equivalent influence diagram. As a corollary, if $t < s$, all those arcs in $I'$ that are not in $I$ are remov­able. Hence, $I$ and $I'$ are equivalent. In other words, if $c$ is in the upstream of $\pi_s$, the value of perfect infor­mation on $c$ at $d_s$ is 0. In the extended oil wildcatter example, it is of no value to acquire perfect knowledge about **seismic-structure** at the time one is to make the **oil-sale-policy**.

From now on, we shall let $I'$ stand for the diagram after the removal of those removable arcs.

## 4.2   TWO ASSUMPTIONS

We assume that $c$ is a root random node, i.e it has no parents. A consequence of this assumption is that if $I$ is smooth, so is $I'$. When $c$ is not a root, one can transform the diagram by a series of arc reversals so that $c$ becomes a root in the resulting diagram. This is very similar to the operation of smoothing mentioned in subsection 3.1. See Zhang (1993) for details.

We also assume that $d_{t-1}$ is a parent for every value node in the section $I(d_{t-1}, d_t)$. The assumption is to assure that if a value node appears in a section $I(d_i, d_{i+1})$ of $I$, then it appears in the corresponding section $I'(d_i, d_{i+1})$ of $I'$. This allows us more chances in making use of the local value functions of the sec-tions of $I$ in computing the local value functions of $I'$, as the reader will see in the following. The assumption is not restrictive because one can always pretend that the value function $f_v$ of a value node $v$ in $I(d_{t-1}, d_t)$ depends on $d_t$ even thought it actually does not.

---

[2]It is always the case if the influence diagram is in the so-called Howard normal form. See Matheson (1990).

Under those two assumptions, we can show that $I'(d_j, d_{j+1})$ is the same as $I(d_j, d_{j+1})$ except that it may contain the extra random node $c$.

## 4.3  SECTIONS BEFORE $d_{s-1}$ AND SECTIONS AFTER $d_t$

We need to consider four cases. Let us first discuss the easiest case: the sections before $d_{s-1}$ and sections after $d_t$.

This case occurs when $i \leq s-2$ or $i \geq t$. In such a case, $I'(d_i, d_{i+1})$ is exactly the same as $I(d_i, d_{i+1})$. So, the conditional probability and the local value function in $I'(d_i, d_{i+1})$ are the same as those in $I(d_i, d_{i+1})$, which can simply be retrieved from the condensation of $I$.

In the extended oil wildcatter example, the terminal section falls into this category.

## 4.4  THE SECTION FROM $d_{s-1}$ TO $d_s$

The section $I'(d_{s-1}, d_s)$ is the same as $I(d_{s-1}, d_s)$, except that it contains one extra node $c$. This node is isolated in $I'(d_{s-1}, d_s)$. Thus $c$ is independent of all the other nodes in $I'(d_{s-1}, d_s)$.

Since $\pi'_{d_{s-1}} = \pi_{d_{s-1}}$ and $\pi'_{d_s} = \pi_{d_s} \cup \{c\}$, the conditional probability $P_{I'(d_{s-1}, d_s)}(\pi'_{d_s} | \pi'_{d_{s-1}}, d_{s-1})$ can be computed by

$$\begin{aligned} P_{I'(d_{s-1}, d_s)}&(\pi'_{d_s} | \pi'_{d_{s-1}}, d_{s-1}) \\ &= P_{I'(d_{s-1}, d_s)}(\pi_{d_s}, c | \pi_{d_{s-1}}, d_{s-1}) \\ &= P_{I'(d_{s-1}, d_s)}(\pi_{d_s} | \pi_{d_{s-1}}, d_{s-1}) P(c) \qquad (5) \\ &= P_{I(d_{s-1}, d_s)}(\pi_{d_s} | \pi_{d_{s-1}}, d_{s-1}) P(c), \qquad (6) \end{aligned}$$

where equation (5) is due to the fact that $c$ is independent of all the other nodes in $I'(d_{s-1}, d_s)$. In equation (6), $P(c)$ is given in the specification of $I$ and $P_{I(d_{s-1}, d_s)}(\pi_{d_s} | \pi_{d_{s-1}}, d_{s-1})$ can be retrieved from the condensation of $I$.

We now turn to the local value function. For any value node $v$ in the section $I(d_{s-1}, d_s)$, we have that $c \notin \pi_v$, because $c$ is in another section. By making use of the fact that $c$ is independent of all the other nodes in $I'(d_{s-1}, d_s)$ again, we get

$$\begin{aligned} P_{I'(d_{s-1}, d_s)}&(\pi'_v | \pi'_{d_{s-1}}, d_{s-1}) \\ &= P_{I'(d_{s-1}, d_s)}(\pi_v | \pi_{d_{s-1}}, d_{s-1}) \\ &= P_{I(d_{s-1}, d_s)}(\pi_v | \pi_{d_{s-1}}, d_{s-1}). \end{aligned}$$

This equation and equations (3, 4) give us the following formula for computing the local value function $f'_{s-1}$ in

the section $I'(d_{s-1}, d_s)$:

$$f'_{s-1}(\pi'_{d_{s-1}}, d_{s-1}) = f_{s-1}(\pi_{d_{s-1}}, d_{s-1}), \qquad (7)$$

where $f_{s-1}(\pi_{d_{s-1}}, d_{s-1})$ can be retrieved from the condensation of $I$.

In the extended oil wildcatter example, the section from test to drill falls into this case.

## 4.5  SECTIONS IN BETWEEN $d_s$ AND $d_{t-1}$

This subsection considers the case when $s \leq i < t-1$. In this case, the section $I'(d_i, d_{i+1})$ is the same as $I(d_i, d_{i+1})$, except that it contains one extra node $c$. This node is isolated in $I'(d_i, d_{i+1})$. So, $c$ is independent of all other nodes in $I'(d_i, d_{i+1})$. Since $\pi'_{d_i} = \pi_{d_i} \cup \{c\}$ and $\pi'_{d_{i+1}} = \pi_{d_{i+1}} \cup \{c\}$, the conditional probability $P_{I'(d_i, d_{i+1})}(\pi'_{d_{i+1}} | \pi'_{d_i}, d_i)$ satisfies

$$\begin{aligned} P_{I'(d_i, d_{i+1})}&(\pi'_{d_{i+1}} | \pi'_{d_i}, d_i) \\ &= P_{I'(d_i, d_{i+1})}(\pi_{d_{i+1}}, c | \pi_{d_i}, c, d_i) \\ &= P_{I'(d_i, d_{i+1})}(\pi_{d_{i+1}} | \pi_{d_i}, d_i) \\ &= P_{I(d_i, d_{i+1})}(\pi_{d_{i+1}} | \pi_{d_i}, d_i), \qquad (8) \end{aligned}$$

where $P_{I(d_i, d_{i+1})}(\pi_{d_{i+1}} | \pi_{d_i}, d_i)$ can be retrieved from the condensation of $I$ since $c \in \pi_{d_i}$.

We now turn to the local value function. For any value node $v$ in the section $I(d_i, d_{i+1})$, we have that $c \notin \pi_v$. By making use of the fact that $c$ is independent of all the other nodes in $I'(d_i, d_{i+1})$ again, we get

$$\begin{aligned} P_{I'(d_i, d_{i+1})}&(\pi'_v | \pi'_{d_i}, d_i) \\ &= P_{I'(d_i, d_{i+1})}(\pi_v | \pi_{d_i}, d_i) = P_{I(d_i, d_{i+1})}(\pi_v | \pi_{d_i}, d_i). \end{aligned}$$

This equation and equations (3, 4) give us the following formula for computing the local value function $f'_i$ in the section $I'(d_i, d_{i+1})$:

$$f'_i(\pi'_{d_i}, d_i) = f_i(\pi_{d_i}, d_i), \qquad (9)$$

where $f_i(\pi_{d_i}, d_i)$ can be retrieved from the condensation of $I$.

## 4.6  THE SECTION FROM $d_{t-1}$ TO $d_t$

This subsection considers the section from $d_{t-1}$ to $d_t$. The section $I'(d_{t-1}, d_t)$ is the same as $I(d_{t-1}, d_t)$, and $\pi'_{d_{t-1}} = \pi_{d_{t-1}} \cup \{c\}$ and $\pi'_{d_t} = \pi_{d_t}$. Thus we have

$$\begin{aligned} P_{I'(d_{t-1}, d_t)}&(\pi'_{d_t} | \pi'_{d_{t-1}}, d_{t-1}) \\ &= P_{I(d_{t-1}, d_t)}(\pi_{d_t} | \pi_{d_{t-1}}, c, d_{t-1}). \end{aligned}$$

If $c \in \pi_{d_t}$, we have

$$\begin{aligned} P_{I(d_{t-1}, d_t)}&(\pi_{d_t} | \pi_{d_{t-1}}, c, d_{t-1}) \\ &= \frac{P_{I(d_{t-1}, d_t)}(\pi_{d_t} | \pi_{d_{t-1}}, d_{t-1})}{\sum_{\pi_{d_{t-1}} - \{c\}} P_{I(d_{t-1}, d_t)}(\pi_{d_t} | \pi_{d_{t-1}}, d_{t-1})}. (10) \end{aligned}$$

Thus, one can use the right hand side of equation (10) to compute $P_{I'(d_{t-1},d_t)}(\pi'_{d_t}|\pi'_{d_{t-1}}, d_{t-1})$ when $c \in \pi_{d_t}$. In the extended oil wildcatter problem, the section from `drill` to `oil-sale-policy` is an example of this case.

On the other hand, if $c \notin \pi_{d_t}$, there is no obvious way to make use of $P_{I(d_{t-1},d_t)}(\pi_{d_t}|\pi_{d_{t-1}}, d_{t-1})$ in computing $P_{I'(d_{t-1},d_t)}(\pi'_{d_t}|\pi'_{d_{t-1}}, d_{t-1})$. In such a case, $P_{I'(d_{t-1},d_t)}(\pi'_{d_t}|\pi'_{d_{t-1}}, d_{t-1})$ needs to be computed from scratch.

Now, the local value function. If for every value node $v$ in the section $I(d_{t-1}, d_t)$, one has that $\pi_v \subseteq \pi_{d_{t-1}} \cup \{d_{t-1}\}$, then it is easy to see that

$$f'_{t-1}(\pi'_{d_{t-1}}, d_{t-1}) = f_{t-1}(\pi_{d_{t-1}}, d_{t-1}).$$

Again in the extended oil wildcatter problem, the section from `drill` to `oil-sale-policy` is an example of this case.

In any other case, we see no way to make use of $f_{t-1}$ in computing $f'_{t-1}$. One needs to compute $f'_{t-1}$ from scratch.

### 4.7  How much computation savings?

To end this section, we would like to give the reader some idea about how much computation our approach can save. There are two SDID's, the original $I$ and the modified $I'$. The savings are in the process of evaluating $I'$. There are two stages: in stage 1 one condenses $I'$, and in stage 2 one evaluates the condensed SDID. As we have shown in this section that it takes very little computation to obtain the condensation of $I'$ from that of $I$. This means a lot of savings if there are many random nodes in $I$ that are not parents of any decision nodes, since those are the nodes that the condensation precess needs to get rid of. As we have pointed out earlier, our approach is especially useful if one wishes to evaluate the value of perfect information for a number cases. One can compute the condensation of $I$ once and use it for all the cases. We can also save some computation in stage 2. Since $I'(d_i, d_{i+1})$ and $I(d_i, d_{i+1})$ are the same for all $i \geq t$, we need not to re-evaluate these sections at all. Furthermore, we can save more if we adopt a top down approach for evaluating the condensed diagrams. See Qi (1993) and Zhang, Qi, and Poole (1993a) for details.

## 5  CONCLUSIONS

The value of perfect information in an influence diagram is defined as the difference between the optimal expected value of a properly modified influence diagram $I'$ and that of the $I$ itself. In this paper, we have described a method for computing the value of perfect information. The method is incremental in the sense that it computes the value of $I'$ by using the intermediate computation results obtained in evaluating $I$. Of fundamental importance to the method is the concept of condensation, which also leads easy implementations of SDID's on top of a system for Bayesian network computations.

### Acknowledgement

This paper is partly supported by NSERC Grant OG-POO44121.